# Supervised classification methods applied to airborne hyperspectral images: Comparative study using mutual information

Hasna Nhaila* , Asma Elmaizi, Elkebir Sarhrouni, Ahmed Hammouch

*Laboratory LRGE, ENSET, Mohammed V University, B.P.6207 Rabat, Morocco*

**Abstract**

Nowadays, the hyperspectral remote sensing imagery HSI becomes an important tool to observe the Earth's surface, detect the climatic changes and many other applications. The classification of HSI is one of the most challenging tasks due to the large amount of spectral information and the presence of redundant and irrelevant bands. Although great progresses have been made on classification techniques, few studies have been done to provide practical guidelines to determine the appropriate classifier for HSI. In this paper, we investigate the performance of four supervised learning algorithms, namely, Support Vector Machines SVM, Random Forest RF, K-Nearest Neighbors KNN and Linear Discriminant Analysis LDA with different kernels in terms of classification accuracies. The experiments have been performed on three real hyperspectral datasets taken from the NASA's Airborne Visible/Infrared Imaging Spectrometer Sensor AVIRIS and the Reflective Optics System Imaging Spectrometer ROSIS sensors. The mutual information had been used to reduce the dimensionality of the used datasets for better classification efficiency. The extensive experiments demonstrate that the SVM classifier with RBF kernel and RF produced statistically better results and seems to be respectively the more suitable as supervised classifiers for the hyperspectral remote sensing images.

## 1. Introduction

In the last decades, the acquisition of images with higher spectral resolution becomes possible using the hyperspectral remote sensing imagery HSI, the large amount of information that contains makes it useful in many applications including environmental studies, military, the study of plant's stress and especially land cover analysis [1] [2]. For HSI classification, several algorithms have been developed in the literature and can be divided into two main groups, namely, supervised and unsupervised methods. The supervised classification techniques require the availability of a subset of the ground truth to use for training whereas in the unsupervised techniques, no prior definitions of the classes are used. In this work, we investigate the reliability of four well-known supervised learning algorithms, namely, Support Vector Machines SVM, Random Forests RF, K-Nearest Neighbors KNN and Linear Discriminant Analysis LDA with different kernels to determine their performance in terms of classification accuracies. The main motivation behind this choice is due to the following two reasons: (i) these methods provide, in the literature, good performances in many applications in different areas and especially in remote sensing [3] [4] [5], (ii) no study has been made to compare the classification performance of these four algorithms to determine the most adapted for the hyperspectral images.

* Corresponding author. Tel.: +2126- 650- 49210
  E-mail address: hasnaa.nhaila@gmail.com

The classification of HSI suffers from many problems for both supervised and unsupervised learning [6]. In the case of supervised classification methods, Huges phenomenon [7] called the curse of dimensionality; complicates the learning system leading to have poor classification model. It is due to the large spectral information with limited number of training samples and the presence of irrelevant and redundant bands. To overcome these challenges, it is necessary to use dimension reduction DR techniques as a pre-processing step of the HSI classification; they consist to transform the image from a high order to a lower dimension by eliminating the irrelevant bands without losing useful information [8] [9] [10]. The DR can be done either by feature extraction or feature selection or by selection followed by extraction. The feature selection based methods are the most commonly used [11]; they can be classified into two categories namely filter or wrapper approaches in terms of dependency of the evaluation step on the classification algorithm. In this study, we will use the filter approach for feature selection based on mutual information to reduce the dimension of the used datasets to getting better classification efficiency.

The rest of this paper is organized as follows: In the next section, we explain the used band selection method for dimensionality reduction based on mutual information and we briefly present the four supervised classifiers retained for this study. Section 3, presents the datasets and discusses the experimental results. Finally, section 4 concludes our work.

## 2. Methodology

To deal with the aforesaid challenges of hyperspectral images classification, this work aims to make two contributions: First, confirm the power and the validity of the mutual information to select the relevant bands as a pre-processing step of HSI classification and improve its accuracy. Second, investigate and compare the performance of the four supervised classifiers in terms of classification efficiency and help to determine the more suitable for HSI classification. The block diagram of this methodology is shown in the following figure 1. The detailed process of this study is presented in this section.

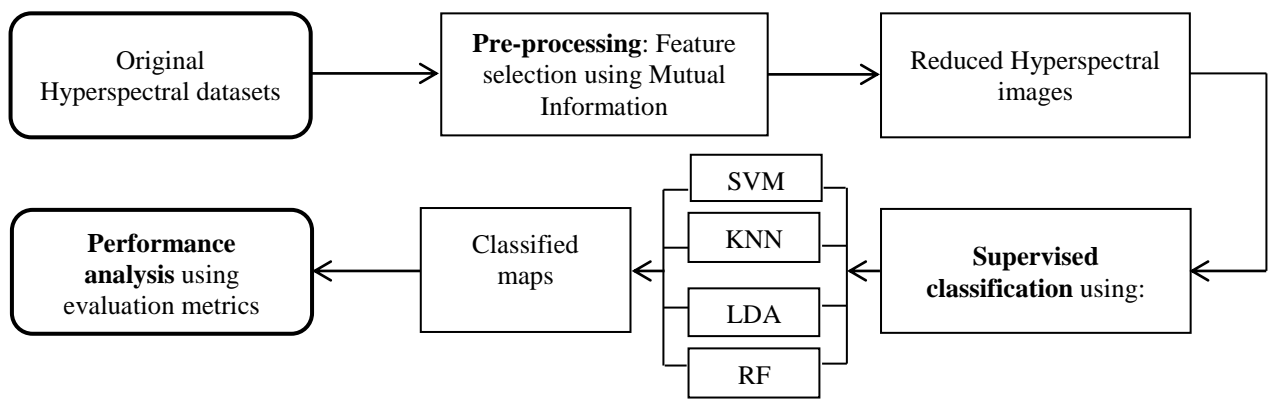

Fig. 1. Block diagram of the methodology

*2.1. Feature selection by mutual information*

The hyperspectral images provide more than hundred bands of the same region, but some of them are redundant and don't contain relevant information. These attributes lead to have a poor classification model. In this study, we will use a filter approach based on mutual information to overcome this problem and select the reduced group of bands in order to improve the classification accuracy.

Within informative theory researches [12], Shannon entropy denoted by H(X) is used in order to measure the quantity of information contained in a random variable X as presented in the equation (1). With p(X) is the probability density function of X.

$$H(X) = \sum_X p(X) \log_2 p(X) \tag{1}$$

The entropy will be used in our case to quantify the amount of information contained in a band (X) of the hypersperctral image. The relation between entropy and the mutual information MI can be formulated as below:

$$I(X,Y) = H(X) + H(Y) - H(X,Y) \tag{2}$$

As defined in (2), the MI calculates the difference between the dependent and joint distributions of the entropy to estimate the statistical dependence of two bands X and Y. The MI is also expressed as:

$$I(X,Y) = \sum_{X,Y} P(X,Y) \log_2 \frac{P(X,Y)}{P(X)p(Y)} \tag{3}$$

With p(X,Y) is the joint probability density function of band X and band Y. So our feature selection algorithm using the mutual information is based on four steps:

- First, compute the MI between the ground truth and each band of the original hyperspectral dataset.
- Second, we initialize the selected bands by the one that have the largest MI with the ground truth.
- Third, an approximated reference map called (G-est) is built by the average of the last one with the candidate band.
- Fourth, The added band is retained if it increases the last value of the MI between the Ground Truth GT and the approximated reference G_est: MI(GT,G-est) otherwise, it will be rejected.

*2.2. Supervised classification methods*

In this subsection, we give a brief review of the principle of the four supervised classification algorithms used in this study. Each technique adopts a learning algorithm to identify a model that fits the relationship between the attributes and class labels of the input hyperspectral images.

*A. Support Vector Machines SVM*

Support vector machines SVM is a supervised machine learning paradigm, it has been widely used in classification of hyperspectral remote sensing images and has provided good results in many works [13][14]. SVM is of two types linear and nonlinear depending on the hyperplane defined for the classification, the nonlinear SVM is performed using kernel function K.

The general principle of the SVM is to find the optimal hyperplane that separates samples belonging in two classes by maximizing the distance between the margins.

In certain cases, linear classifier fails to find an optimal hyperplane, this forced us to use nonlinear type, and in this case, the data are mapped into a higher dimensional space using Kernel function which must fulfill Mercer's conditions. Refer to [15] for more details. The kernels used in this study are linear, radial basis function RBF and sigmoid.

*B. K-Nearest Neighbor K-NN*

K-Nearest Neighbor KNN is a non-parametric learning algorithm; it is one of the most useful as supervised classifier that keeps all the training data to make decision based on similarity measure. It has been successfully applied on hyperspectral images classification [16].

Let $X_{Tr}$ be the labled training data of n points and $X$ are the new unlabeled points. $X_{Tr} = \{(x_1, y_1), (x_2, y_2), \ldots \ldots \ldots \ldots, (x_n, y_n)\}$. The general procedure of the KNN classifier may be summarized in these steps:

- Find the distance between every point in the training data $x_i$ and the new point X. In this step, various metrics can be used to determine this distance such as: Euclidean, standardized Euclidean, mahalanobis, cityblock, chebychev distance etc. The popular one is the Euclidean distance that we use in this study.
- Sort these distances in ascending order.
- Return the k points in $x_i$ that are closest to the new point X.

For $k = 1$: The case of nearest neighbor algorithm 1NN, the new point gets the class label of the nearest neighbor.
For $k > 1$: The case of K-Nearest Neighbors KNN, the new point is classified by voting the most frequent neighbor.

The choice of optimal value of k is critical. In general, the accuracy value increases with a large value of k because it reduces the overall noise but the computational cost and time also increase. In this study we perform the KNN classifier using different values of k: 1, 3, 5 and 7.

*C. Linear Discriminant analysis LDA*

Linear discriminant analysis LDA is also known as the Fisher discriminant analysis named for its developer R.A. Fisher. It is sample but gives good models as more complex methods. When applied on hyperspectral images classification, this algorithm aims to find a linear combination of features to separate the classes of the dataset. Indeed, this approach maximizes the between class variance to the within class variance ratio to ensure maximum separability. For a full theoretical description, the reader is referred to [17]. In this study two types of LDA were used which are linear and diaglinear:

- Linear: (default) estimates one covariance matrix for all classes.
- Diag-linear: uses the diagonal of the linear covariance matrix.

*D. Random Forests RF*

Random forests RF [18] is one of the ensemble learning algorithms popularly used in many kinds of data science problems [19]. As its name suggests, this classifier consists to construct a multitude of decision tree DCT for training. Its main idea is that a group of "weak learners" can come together to form a "strong learner".

The RF is a combination of one of the supervised classifiers called "decision tree" which corresponds to our weak learner. In decision tree, we have high variance and high bias but RF overcomes these problems and creates a balance between these two errors.

The general steps for performing a random forest are listed below:
- Build the random forest:

- Select "*m*" features from the total features "*F*".
- For $m < F$: Calculate the nodes "n" and their daughters using the best split point.
- Repeat these two steps until having "*d*" nodes and the target as the leaf node.
- Repeat the aforesaid steps until having a forest with "*k*" trees.

- Random forest prediction:

- Store the predicted results using the created decision trees and the best features.
- Use the majority voting for each predicted target.
- The final prediction of the RF algorithm is the high voted predicted target.

## 3. Experiments results and discussion

*3.1. Datasets description*

For experiments, three datasets will be used in this study, from two types of airborne hyperspectral sensors which are publicly available at [20]. These datasets have different characteristics in terms of number of bands and classes and features type.

*A. Indian Pines*

The first dataset used in this study is acquired by the 224-band Airborne Visible/Infrared Imaging Spectrometer Sensor AVIRIS over the Indian Pines in North-western Indiana. It has 145x145 pixels and 224 bands in the wavelength range of 0.4-2.5 µm with spatial resolution of 20 m pixels. This scene is widely used in many works related to HSI analysis. It includes sixteen classes. The Color composite and the corresponding ground truth reference of this dataset are presented in Fig. 2a.

*B. Salinas*

The second dataset is Salinas. It is captured by the 224-band AVIRIS over Salinas valley, CA, USA. It consists of 217x512 pixels and 224 spectral reflectance bands in the wavelength range of 0.4 to 2.5 µm. Salinas scene is characterized by high spatial resolution (3.7 m pixels). The color composite and the corresponding ground truth reference are given in Fig. 2b. It contains also sixteen classes and it is known by its complicated classification scenario due to their highly mixed pixels.

*C. University of Pavia*

The University of Pavia dataset is a 610x340 pixels scene gathered by the Reflective Optics System Imaging Spectrometer (ROSIS-03) sensor over urban area of engineering school at University of Pavia, Italy. Its spatial resolution is 1.3 m per pixel. Original dataset has 115 spectral bands in the range 0.43-0.86 µm where 12 bands were removed due to the noise. The color composite and the corresponding ground truth reference are shown in Fig. 2c. It includes nine classes.

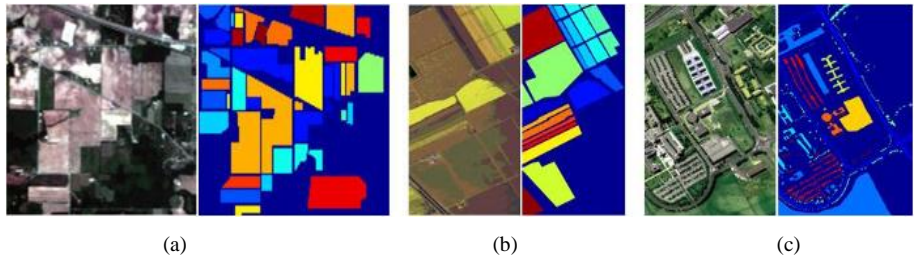

(a)      (b)      (c)

Fig. 2.The three-band color composite and the corresponding Ground Truth map of the hyperspectral images; (a) Indian Pines; (b) Salinas and (c) University of Pavia

*3.2. Classification and evaluation metrics*

To classify the above mentioned datasets using the four supervised classifiers, 50% of the pixels of each class are randomly chosen as training data and 50% for the test. The selected bands using the mutual information are used as input data for the classifiers. The parameters values of the classifiers are experimentally chosen using the cross validation algorithm. All the tests are implemented using the scientific programming language MATLAB.

In order to compare the performance of the classifiers, different coefficients of evaluation can be calculated from the Confusion Matrix. Among the most used in the literature, we have chosen sensitivity, specificity and precision calculated using the following formulas:

$$Sensitivity = \frac{TP}{TP+FN} \qquad (4)$$

$$Specificity = \frac{TN}{TN+FP} \qquad (5)$$

$$Precision = \frac{TP}{TP+FP} \qquad (6)$$

Where TP is true positive, TN is true negative, FP is false positive and FN is false negative.

Two other popular measures widely used in hyperspectral remote sensing imagery for comparing different classifiers were used in this study, namely the overall accuracy OA and the kappa coefficient k. The OA is calculated as the ratio of total number of correctly classified pixels to the total number of test pixels, where the kappa coefficient is used to measure the agreement between classified and truth values. The computational time is also used

*3.3. Results and discussion*

The first step in this study was to select the relevant bands from the datasets which contain 224 bands for Indian Pines and Salinas and 103 bands for Pavia University. This step was done using a filter method based on mutual information. Subsequently, we applied the four supervised classifiers with different kernels on the selected features.

The experimental results using the three datasets are summarized and respectively shown in Fig. 3a, Fig.3b and Fig.3c. From these figures, it is seen that most of the classification algorithms perform well and give good accuracy rate that exceeds 60% with just 30 selected bands. It is also clear that the SVM with RBF kernel outperforms the other methods in terms of classification accuracy that achieves 87.28% for Indian Pines, 93.2% for Salinas and 91.65% for Pavia University, followed by RF and KNN. LDA-linear does not perform as well as SVM-RBF, SVM-linear, RF and KNN with different values of k, but produces better accuracy than sigmoid kernel of SVM and LDA-diaglinear which perform poorly for the three datasets.

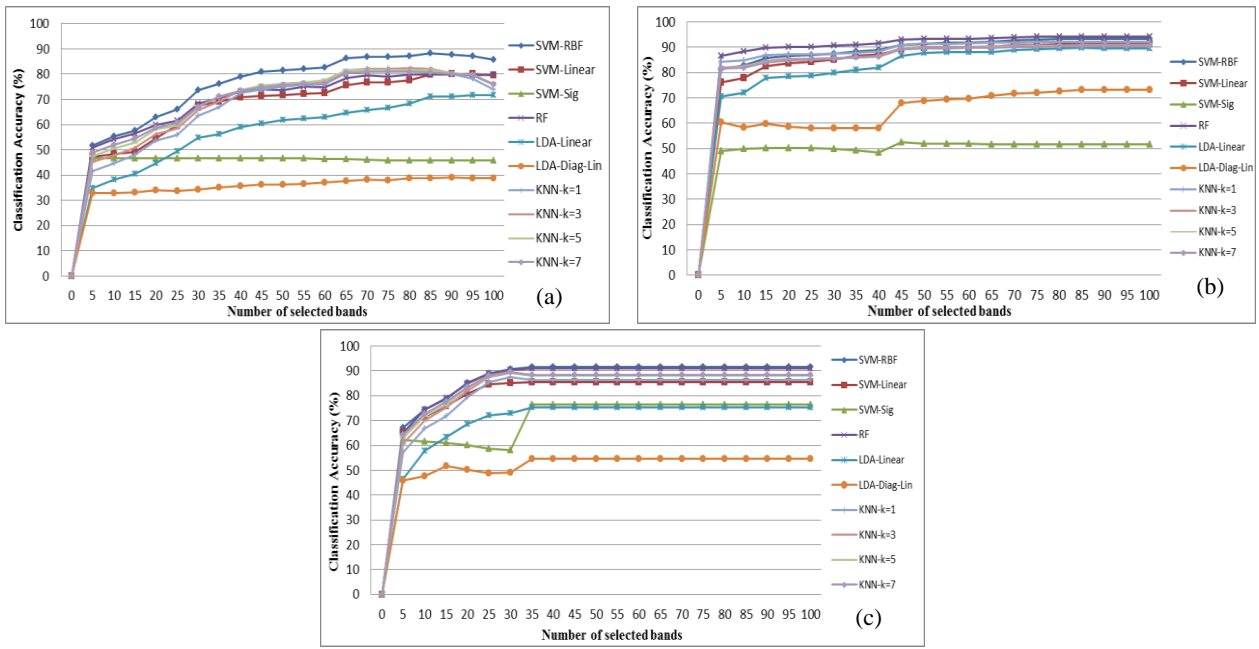

Fig. 3. Comparative performances in terms of classification accuracy and bands selection for (a) Indian Pines; (b) Salinas and (c) University of Pavia

From these figures, we can also observe that the accuracy rate values show a little change beyond 80 selected bands for Indian and Salinas because the most informative bands are already selected and more added bands may decrease the accuracy rate, see Fig. 3a and Fig. 3b. For Pavia University also we remark a constant behavior for all the learning algorithms up to just 35 selected bands since it contains 103 bands which proves the benefit of dimension reduction using mutual information on hyperspectral images, see Fig. 3c.

In order to compare the performances of the different classifiers, a variety of evaluation metrics is calculated and is presented in Tables 1, 2 and 3.

For Indian Pines dataset, it is verified that the best classification performances are obtained by the SVM-RBF classifier for 80 selected bands where we get 93.06% in sensitivity, 99.51% in specificity, 94.42% in precision, 93.27% in OA and kappa coefficient of 0.9282%. For KNN-1 and RF also provide good performance as can be seen in Table 1 where for example the kappa coefficient is respectively 0.9045% and 0.8910% and precision of respectively 91.33% and 91.68%. LDA classifier with linear and diag-linear kernels performs poorly compared to the other methods as show the results in Table 1. Concerning the computational time, we can see that The SVM-RBF and RF need lower time to provide the highest accuracies.

Table 1. Performances comparison of supervised classification methods using different evaluation metrics in Indian Pines dataset

|     |     | Sensitivity (%) | Specificity (%) | Precision (%) | OA (%) | Kappa | Time(s) |
|-----|-----|---|---|---|---|---|---|
| SVM | RBF | 93.06 | 99.51 | 94.42 | 93.27 | 0.9282 | 34.17 |
|     | Linear | 80.35 | 98.44 | 85.39 | 79.01 | 0.7761 | 37.80 |
|     | Sigmoid | 22.66 | 95.75 | 20.01 | 45.72 | 0.4210 | 50.34 |
| RF  |     | 85.01 | 99.25 | 91.68 | 89.78 | 0.8910 | 48.44 |
| DA  | Linear | 77.86 | 97.87 | 69.55 | 69.49 | 0.6745 | 2.86 |
|     | Diag-linear | 44.79 | 95.76 | 36.58 | 38.90 | 0.3483 | 1.13 |
| KNN | K=1 | 90.32 | 99.35 | 91.33 | 91.05 | 0.9045 | 78.37 |
|     | K=3 | 84.78 | 99.09 | 89.43 | 87.52 | 0.8669 | 80.55 |
|     | K=5 | 80.87 | 98.93 | 87.64 | 85.27 | 0.8428 | 81.89 |
|     | K=7 | 77.90 | 98.81 | 85.82 | 83.68 | 0.8260 | 84.53 |

The ground truth and the best classified maps of Indian Pines for each classifier by considering the different kernels are shown in Fig. 4a, 4b, 4c, 4d and 4e. By visual inspection, the SVM-RBF gives the best classified map Fig. 4b and on the other hand, the worst one is obtained using the LDA classifier as illustrated in Fig. 4e. which confirm the results in table 1.

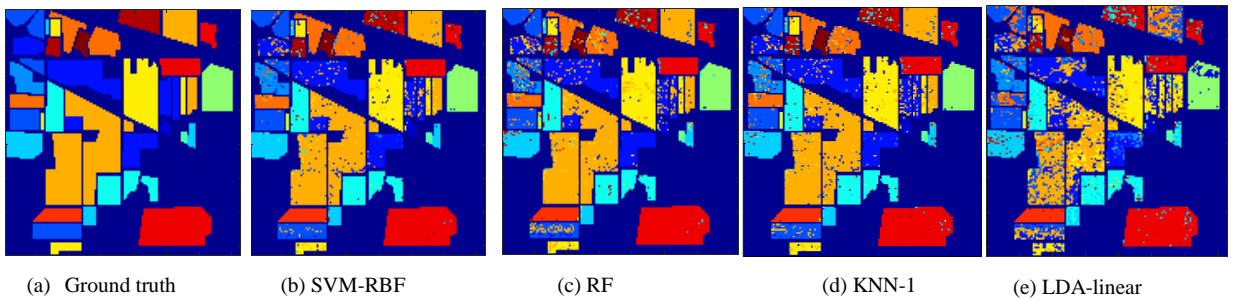

(a) Ground truth  (b) SVM-RBF  (c) RF  (d) KNN-1  (e) LDA-linear

Fig. 4. The Ground truth of Indian Pines and the best classified maps for 80 selected bands using SVM, RF, KNN and LDA

In the same way, we trained our classifiers on the dataset 2 (Salinas) as can be seen in Table 2. The results indicate the better performance of SVM-RBF and RF compared to the other methods, in a good timing, especially LDA-diaglinear and SVM-sigmoid that performs poorly with overall accuracy of 51,52% against 93.44% for SVM-RBF and 91.30% for SVM-linear. KNN with different values of k also performs well and gives good results but not as well as RF and SVM-RBF, see Table 2.

Table 2. Performances comparison of supervised classification methods using different evaluation metrics in Salinas dataset

|     |       | Sensitivity (%) | Specificity (%) | Precision (%) | OA (%) | Kappa  | Time(s) |
|-----|-------|-----------------|-----------------|---------------|--------|--------|---------|
| SVM | RBF   | 97.12           | 99.50           | 97.41         | 93.44  | 0.9300 | 273.28  |
|     | Linear| 95.68           | 99.34           | 96.15         | 91.30  | 0.9072 | 354.14  |
|     | Sigmoid | 31.74         | 96.50           | 27.95         | 51.52  | 0.4829 | 1109.70 |
| RF  |       | 98.56           | 99.78           | 98.60         | 97.09  | 0.9690 | 256.86  |
| DA  | Linear| 93.87           | 99.23           | 93.26         | 89.56  | 0.8886 | 18.04   |
|     | Diag-linear | 78.05     | 98.09           | 69.22         | 72.68  | 0.7085 | 5.15    |
| KNN | K=1   | 98.28           | 99.71           | 98.24         | 96.07  | 0.9580 | 2032.16 |
|     | K=3   | 96.48           | 99.44           | 96.41         | 92.50  | 0.9200 | 2190.36 |
|     | K=5   | 96.00           | 99.39           | 95.99         | 91.89  | 0.9135 | 2485.35 |
|     | K=7   | 95.56           | 99.35           | 95.50         | 91.28  | 0.9070 | 2519.85 |

Again the ground truth and the best classified map of Salinas for each classifier are shown in Fig. 5a, 5b, 5c, 5d and 5e. The least good classified map is obtained using LDA-linear method with OA of 89.56 % and kappa coefficient equal to 0.8886%.

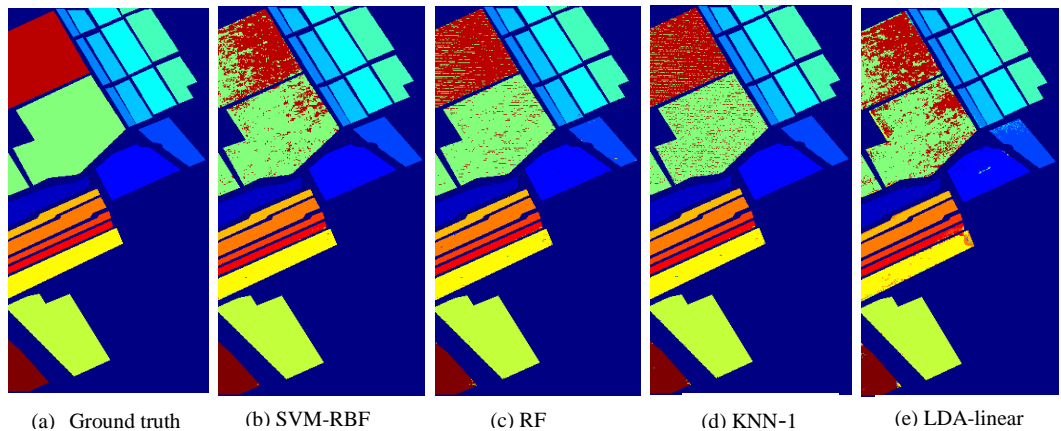

(a) Ground truth    (b) SVM-RBF    (c) RF    (d) KNN-1    (e) LDA-linear

Fig. 5. The Ground truth of Salinas and the best classified maps of 80 selected bands using SVM, RF, KNN and LDA

Finally, the same experiments are performed for dataset 3. Fig. 6 and Table 3 summarize the obtained results using different classifiers and kernels with variety of metric distances. From these results, it is obvious that SVM-RBF and RF classifiers provide good performances of classification with advantage in terms of running time. The KNN also gives good results but it requires a high value of execution time. The LDA on the other hand is the faster but it performs poorly.

Fig. 6a, 6b, 6c, 6d and 6e show the Ground truth of University of Pavia and the best classified maps. As before, for dataset 1 and 2, the top classified map is provided by SVM-RBF (see Fig.6) whereas the least good one is obtained using LDA-linear.

Table 3. Performances comparison of supervised classification methods using different evaluation metrics in Pavia University dataset

|  |  | Sensitivity (%) | Specificity (%) | Precision (%) | OA (%) | Kappa | Time(s) |
|---|---|---|---|---|---|---|---|
| SVM | RBF | 88.59 | 98.84 | 90.07 | 91.91 | 0.9090 | 262.52 |
|  | Linear | 75.17 | 97.80 | 68.74 | 85.48 | 0.8367 | 291.08 |
|  | Sigmoid | 64.78 | 96.36 | 59.46 | 76.38 | 0.7343 | 529.15 |
| RF |  | 93.83 | 99.32 | 95.35 | 95.50 | 0.9493 | 269.33 |
| DA | Linear | 78.92 | 96.84 | 73.62 | 75.47 | 0.7241 | 8.12 |
|  | Diag-linear | 65.46 | 94.16 | 60.42 | 54.76 | 0.4911 | 3.5 |
| KNN | K=1 | 91.97 | 99.01 | 91.75 | 93.19 | 0.9234 | 3435.95 |
|  | K=3 | 88.41 | 98.61 | 89.51 | 90.79 | 0.8964 | 3521.30 |
|  | K=5 | 86.98 | 98.46 | 88.75 | 89.87 | 0.8860 | 3714.08 |
|  | K=7 | 86.40 | 98.39 | 88.62 | 89.51 | 0.8819 | 3886.08 |

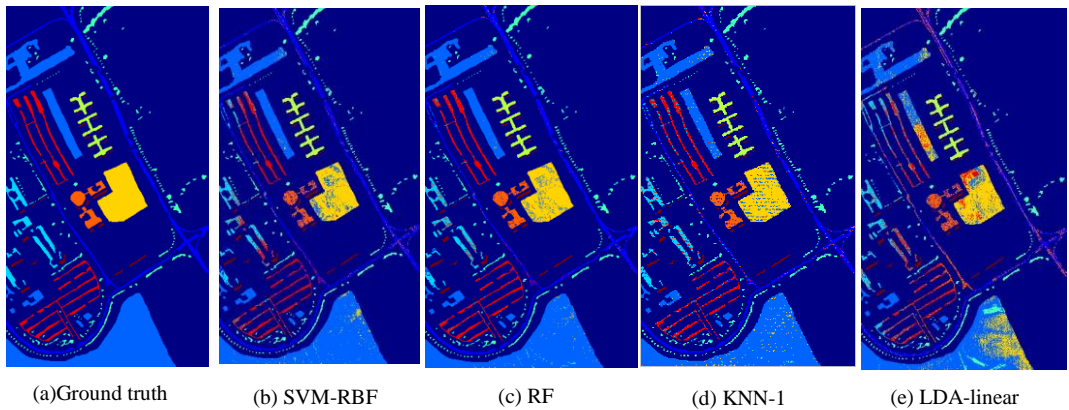

(a) Ground truth    (b) SVM-RBF    (c) RF    (d) KNN-1    (e) LDA-linear

Fig. 6. The Ground truth of University of Pavia and the best classified maps of 80 selected bands using SVM, RF, KNN and LDA

**Partial conclusion**

To summarize the presented experimental results shown in (Fig. 3, 4, 5 and 6) and (Tables 1, 2 and 3); First, it is clear that the feature selection based on mutual information had good effect to get better performance using the different classification methods from 30 to 80 selected bands for Indian Pines and Salinas datasets where the accuracy rate exceeds 94%. For Pavia University, we achieve 91.65 with just 35 selected bands. On the other hand, it is obvious that the best classifier is SVM with RBF kernel, followed by RF and KNN (k=1) against the other algorithms especially LDA with diag-linear kernel that provides the lower results in the three hyperspectral datasets. From the inspection of the computational time, it is seen that the SVM-RBF and RF need a reduced time to provide high performances.

**4. Conclusion**

In this paper, a comparative study of four supervised classifiers has been presented. The classification methods used are Support Vector Machine SVM with (RBF, linear and sigmoid) kernels, Random Forest RF, Discriminant Analysis DA with (linear and diag-linear) kernels, and K-Nearest Neighbors with (k=1, 3, 5 and 7). On the other hand, to address Huges phenomenon and reduce the dimensionality of the used datasets, a filter method based on mutual information have been used to select the more informative bands from the used hyperspectral datasets.

The algorithms have been evaluated using three hyperspectral remote sensing datasets from the NASA's AVIRIS and ROSIS airborne hyperspectral sensors, these datasets had different characteristics in terms of features type and number of bands and classes.

The experimental results confirm the effectiveness of dimension reduction as a pre-processing step of classification using mutual information; also the tests show that the SVM with RBF kernel presents the high performance and seems to be the more effective as supervised classifier for hyperspectral images classification followed by RF in comparison with the other cases.

For the performances evaluation of each classifier, several metrics have been calculated which are sensitivity, specificity, precision, overall accuracy OA, kappa coefficient k and the computational time. All of them affirm that the SVM-RBF provided the maximum performances and the LDA is the least good that performs poorly in comparison with the other methods as it is shown in the classified maps.

Our future objective is to make further investigation in this topic to improve the present results using the unsupervised classifiers and dimensionality reduction methods by both feature selection and extraction. This article can be a guide or reference for the new researchers to choice the adequate classifier of Hyperspectral images as needed.

**Acknowledgements**

The authors are very grateful to the anonymous reviewers of this paper for their helpful remarks and all the participants involved in this study.